\documentclass[10pt,twocolumn,letterpaper]{article}

\usepackage{cvpr}
\usepackage{times}
\usepackage{epsfig}
\usepackage{graphicx}
\usepackage{amsmath}
\usepackage{amssymb}
\usepackage{cite}
\usepackage{booktabs,subcaption,amsfonts,dcolumn}
\usepackage{multicol}
\usepackage{multirow}

\usepackage[pagebackref=true,breaklinks=true,letterpaper=true,colorlinks,bookmarks=false]{hyperref}

\cvprfinalcopy 

\def\cvprPaperID{2832} 

\ifcvprfinal\fi
\begin{document}

\title{Generalizing Deep Models for Overhead Image Segmentation Through Getis-Ord Gi* Pooling}

\author{Xueqing Deng\\
UC Merced\\
{\tt\small xdeng7@ucmerced.edu}
\and
Yi Zhu\\
Amazon\\
{\tt\small yzhu25@ucmerced.edu}
\and
Yuxin Tian\\
UC Merced\\
{\tt\small ytian8@ucmerced.edu}
\and
Shawn Newsam\\
UC Merced\\
{\tt\small snewsam@ucmerced.edu}
}


%


\maketitle

\begin{abstract}

That most deep learning models are purely data driven is both a strength and a weakness. Given sufficient training data, the optimal model for a particular problem can be learned. However, this is usually not the case and so instead the model is either learned from scratch from a limited amount of training data or pre-trained on a different problem and then fine-tuned. Both of these situations are potentially suboptimal and limit the generalizability of the model. Inspired by this, we investigate methods to inform or guide deep learning models for geospatial image analysis to increase their performance when a limited amount of training data is available or when they are applied to scenarios other than which they were trained on. In particular, we exploit the fact that there are certain fundamental rules as to how things are distributed on the surface of the Earth and these rules do not vary substantially between locations. Based on this, we develop a novel feature pooling method for convolutional neural networks using Getis-Ord $G_{i}^{*}$ analysis from geostatistics. Experimental results show our proposed pooling function has significantly better generalization performance compared to a standard data-driven approach when applied to overhead image segmentation.

\end{abstract}

\section{Introduction}

Research in remote sensing has been steadily increasing since it is an important source for Earth observation. Overhead imagery can easily be acquired using low-cost drones and no longer requires access to expensive high-resolution satellite or airborne platforms. Since the data provides convenient and large-scale coverage, people are using it for a number of societally important problems such as traffic monitoring \cite{Ma_2017_traffic}, urban planning \cite{Ball_2017_survey}, vehicle detection \cite{Chen_2014_car}, land cover segmentation \cite{Kussul_2017_LULC}, building extraction \cite{Yuan_2017_building}, etc.


Recently, the analysis of overhead imagery has benefited greatly from deep learning thanks to the significant advancements made by the computer vision community on regular (non-overhead) images. However, there still often remains challenges when adapting these deep learning techniques to overhead image analysis, such as the limited availability of labeled overhead imagery, the difficulty of the models to generalize between locations, etc.

\begin{figure}
    \centering
    \includegraphics[width=\linewidth]{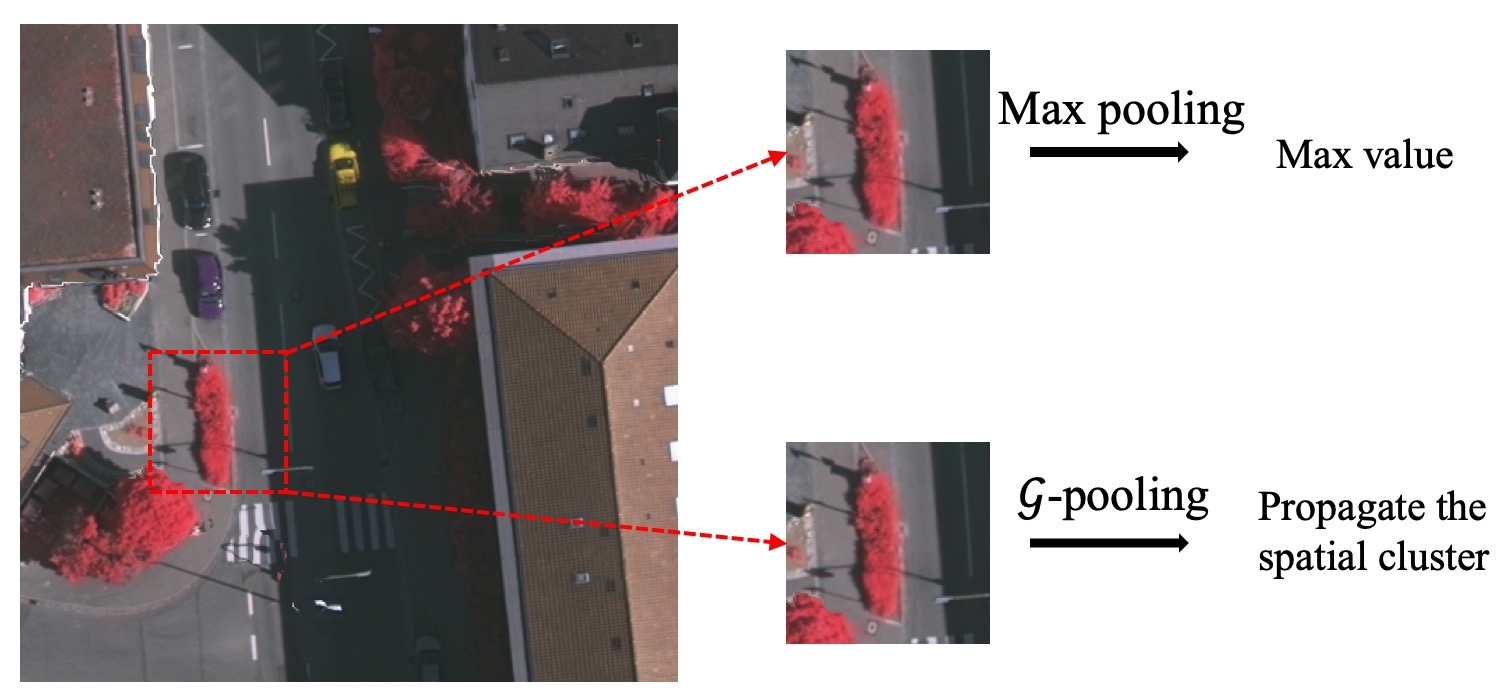}
    \caption{Motivation of our work. The content in the current sliding window is a cluster of pixels of tree. We propose to incorporate geospatial knowledge to build a pooling function which can propagate such a spatial cluster during training, while the standard pooling is not able to achieve it.}
    \label{fig:motivation}
\end{figure}

Annotating overhead imagery is labor intensive so existing datasets are often not large enough to train effective convolutional neural networks (CNNs) from scratch. A common practice therefore is to fine-tune an ImageNet pre-trained model on a small amount of annotated overhead imagery. However, the generalization capability of fine-tuned models is limited as models trained on one location may not work well on others. This is known as the \textit{cross-location generalization} problem and is not necessarily limited to overhead image analysis as it can also be a challenge for ground-level imagery such as cross-city road scene segmentation \cite{chen2017iccv}. Deep models are often overfitting due to their large capacity yet generalization is particularly important for overhead images since they can look quite different due to variations in the seasons, position of the sun, location variation, etc. For regular image analysis, two widely adopted approaches to overcome these so-called domain gaps include domain adaptation \cite{hoffman2018cycada,tzeng2017adversarial,tsai2018learning,tsai2019domain,hoffman2016fcns} and data fusion. Both approaches have been adapted by the remote sensing community \cite{nicolas2018isprs} to improve performance and robustness.


 In this paper, we take a different, novel approach to address the domain gap problem. We exploit the fact that things are not laid out at random on the surface of the Earth and that this structure does not vary substantially between locations.  In particular, we pose the question of how prior knowledge of this structure or, more interestingly, how the fundamental rules of geography might be incorporated into general CNN frameworks. Inspired by work on physics-guided neural networks \cite{karpatne2017physics}, we develop a framework in which spatial hotspot analysis informs the feature map pooling. We term this geo-constrained pooling strategy \textit{Getis-Ord $G_{i}^{*}$ pooling} and show that it significantly improves the semantic segmentation of overhead imagery particularly in cross-location scenarios. To our knowledge, ours is the first work to incorporate geo-spatial knowledge directly into the fundamental mechanisms of CNNs. A brief overview of our motivation is shown in Figure \ref{fig:motivation}. 


Our contributions are summarized as follows: 

(1) We propose Getis-Ord $G_{i}^{*}$ pooling, a novel pooling method based on spatial Getis-Ord $G_{i}^{*}$ analysis of CNN feature maps. Getis-Ord $G_{i}^{*}$ pooling is shown to significantly improve model generalization for overhead image segmentation.

(2) We establish more generally that using geospatial knowledge in the design of CNNs can improve the generalizability of models which provides the simulated process of the data.


\section{Related Work}
\vspace{-15pt}
\noindent\paragraph{Semantic segmentation} Fully connected neural networks (FCN) were recently proposed to improve the semantic segmentation of non-overhead imagery \cite{long2015fully}. Various techniques have been proposed to boost their performance, such as atrous convolution \cite{chen2017deeplabv2,chen2018deeplabv3,chen2018deeplabv3plus,zhao2017pspnet}, skip connections \cite{unet2015}, and preserving max pooling index for unpooling \cite{segnet2017pami}. And recently, video is used to scale up training sets by synthesizing new training samples which is able to improve the accuracy of semantic segmentation networks\cite{Zhu2019CVPR}. Remote sensing research has been driven largely by adapting advances in regular image analysis to overhead imagery. In particular, deep learning approaches to overhead image analysis have become a standard practice for a variety of tasks, such as land use/land cover classification \cite{Kussul_2017_LULC}, building extraction \cite{Yuan_2017_building}, road segmentation \cite{Mnih_2010_eccv_road}, car detection \cite{Chen_2014_car}, etc. More literature can be found in a recent survey \cite{Zhu_2017_remote_sensing_survey}. And various segmentation networks have been proposed, such relation-augmentation networks \cite{mou2019relation} and casnet \cite{casnet2018isprs}. However, these methods only adapt deep learning techniques and networks from regular to overhead images--they do not incorporate geographic structure or knowledge.


\vspace{-15pt}
\noindent\paragraph{Knowledge guided neural networks}Analyzing overhead imagery is not just a computer vision problem since principles of the physical world such as geo-spatial relationships can help. For example, knowing the road map of a city can definitely improve tasks like building extraction or land cover segmentation. While there are no works directly related to ours, there have been some initial attempts to incorporate geographic knowledge into deep learning \cite{Chen_2017_golf,Zhang_2016_airport}. Chen et al. \cite{Chen_2017_golf} develop a knowledge-guided golf course detection approach using a CNN fine-tuned on temporally augmented data. They also apply area-based rules during a post-processing step. Zhang et al. \cite{Zhang_2016_airport} propose searching for adjacent parallel line segments as prior spatial information for the fast detection of runways. However, these methods simply fuse prior knowledge from other sources. Our proposed method is novel in that we incorporate geo-spatial rules into the CNN mechanics. We show later how this helps regularize the model learning and leads to better generalization.

\vspace{-15pt}
\paragraph{Pooling functions} There are various studies in pooling for image classification as well as segmentation. $L_p$ norm is proposed to extend max pooling where intermediate pooling functions are manually selected between max and average pooling to better fit the distribution of the input data. \cite{lee2016gated} generalizes pooling methods by using a learned linear combination of max and average pooling. Detail-Preserving Pooling (DPP) \cite{saeedan2018ddp} learns weighted summations of pixels over different pooling regions. Salient pixels are more importance in order to achieve higher visual satisfaction.  Stride convolution is used toreplace all max pooling layers and activation functions in a small classification model that is trained from scratch and achieve better performance\cite{springenberg2015striving}. However, stride convolutions are common in segmentation tasks. For example, the DeepLab series of networks \cite{chen2018deeplabv3,chen2018deeplabv3plus} use stride convolutional layers for feature down-sampling rather than max pooling. To enhance detail preservation in segmentation, a recent polynomial pooling approach is proposed in \cite{p-pooling2019cvpr}. However, all these pooling methods are based on non-spatial statistics. We instead incorporate geo-spatial rules/simulation to perform the downsampling.




\section{Methods}
\vspace{-5pt}
\label{sec:methods}
In this section, we investigate how geo-spatial knowledge can be incorporated into standard deep CNNs.  We discuss some general rules from geography to describe geo-spatial patterns on the Earth. Then we propose using Getis-Ord $G_{i}^{*}$ analysis, a common technique for geo-spatial clustering, to encapsulate these rules. This then informs our pooling function which is very general and can be used in many network architectures.   

\begin{figure}
    \centering
    \includegraphics[width=\linewidth]{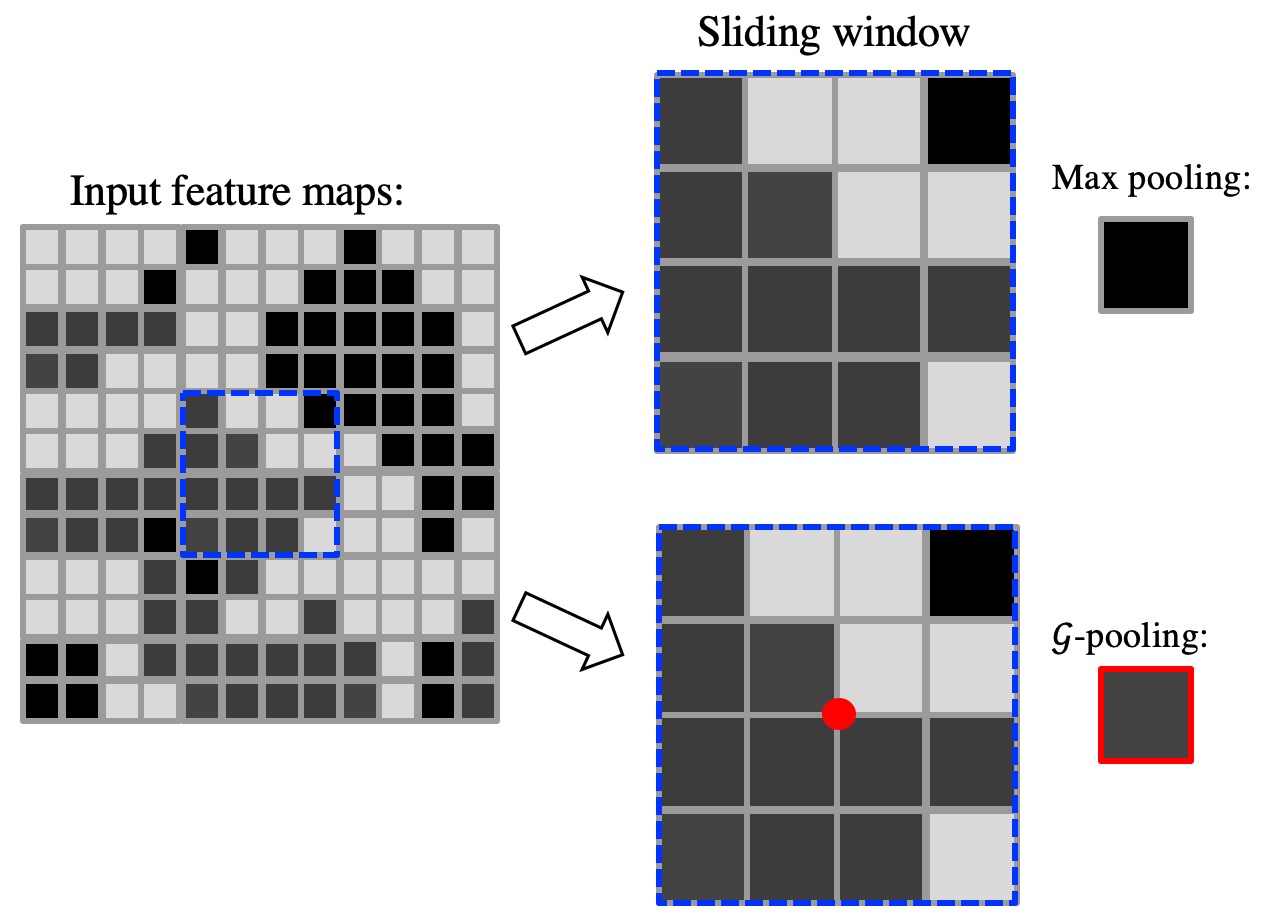}
    \caption{Given a feature map as an input, max pooling (top right) and the proposed $\mathcal{G}$-pooling (bottom right) create different output downsampled feature map based on the characteristics of spatial cluster. The feature map within the sliding window (blue dot line) indicates a spatial cluster. Max pooling takes the max value ignoring the spatial cluster, while our $\mathcal{G}$-pooling takes the interpolated value at the center location. (White, gray and black represent three values range from low to high.) }
    \label{fig:pooling}
\end{figure}

\subsection{Getis-Ord $G_{i}^{*}$ pooling  ($\mathcal{G}$-pooling)}


We take inspiration from the well-known first law of geography: \textit{everything is related to everything else, but near things are more related than distant things}\cite{tobler1970law}. While this rule is very general and abstract, it motivates a number of quantitative frameworks that have been shown to improve geospatial data analysis. For example, it motivates spatial autocorrelation which is the basis for spatial prediction models like kriging. It also motivates the notion of spatial clustering wherein similar things that are spatially nearby are more significant than isolated things. Our proposed framework exploits this to introduce a novel feature pooling method which we term Getis-Ord $G_{i}^{*}$ pooling.

Pooling is used to spatially downsample the feature maps in deep CNNs. In contrast to standard image downsampling methods which seek to preserve the spatial envelope of pixel values, pooling selects feature values that are more significant in some sense. The most standard pooling method is max pooling in which the maximum feature value in a window is propagated. Other pooling methods have been proposed. Average pooling is an obvious choice and is used in \cite{zagoruyko2016wide, huang2017densenet} for image classification. Strided convolution \cite{kuen2017delugenets} has also been used. However, max pooling remains by far the most common as it has the intuitive appeal of extracting the maximum activation and thus the most prominent features of an image.

However, we postulate that isolated high feature values might not be the most informative and instead develop a method to propagate clustered values. Specifically, we use a technique from geostatistics termed hotspot analysis to identify clusters of large values and then propagate a representative from these clusters. Hotspot analysis uses the Getis-Ord $G_{i}^{*}$ \cite{gtis1995hotspot} statistic to find locations that have either high or low values and are surrounded by locations also with high or low values. These locations are the so-called hotspots. The Getis-Ord $G_{i}^{*}$ statistic is computed by comparing the local sum of a feature and its neighbors proportionally to the sum of all features in a spatial region. When the local sum is different from the expected local sum, and when that difference is too large to be the result of random noise, it will lead to a high positive or low negative $G_{i}^{*}$ value that is statistically significant. We focus on locations with high positive $G_{i}^{*}$ values since we want to propagate activations.

\subsection{Definition}
\label{sec:defination}
We now describe our $\mathcal{G}$-pooling algorithm in detail. Please see Figure \ref{fig:pooling} for reference. Similar to other pooling methods, we use a stride sliding window to downsample the input. Given a feature map within the stride window, in order to compute its $G_{i}^{*}$, we first need to define the weight matrix based on the spatial locations. 

We denote the feature values within the sliding window as $\mathbf{X}={x_1,x_2,...,x_n}$ where $n$ is the number of pixels (locations) within the sliding window. We assume the window is rectangular and compute the $G_{i}^{*}$ statistic at the center of the window. Let the feature value at the center be $x_i$. (If the center does not fall on a pixel location then we compute $x_i$ as the average of the adjacent values.) The $G_{i}^{*}$ statistic uses weighed averages where the weights are based on spatial distances. Let $p^x(x_j)$ and $p^y(x_j)$ denote the x and y positions of feature value $x_j$ in the image plane. A weight matrix $w$ that measures the Euclidean distance on the image plane between $x_i$ and the other locations within the sliding window is then computed as

\begin{equation}
    w_{i,j}=\sqrt{(p^x(x_i) -p^x(x_j))^2+(p^y(x_i)-p^y(x_j))^2}.
\end{equation}

\noindent The Getis-Ord $G_{i}^{*}$ value at location $i$ is now computed as
\begin{equation}
\label{eq_Gi}
    G_i^*=\frac{\sum_{j=1}^{n}{w_{i,j}} x_j-\bar{X}\sum_{j=1}^{n}{w_{i,j}}}{S\sqrt{\frac{[n\sum_{j=1}^{n}{w_{i,j}^2-(\sum_{j=1}^{n}w_{i,j})^2]}}{n-1}}}.
\end{equation}
where $\bar{X}$ and $S$ are as below,
\begin{equation}
\label{eq_x_bar}
    \bar{X}=\frac{\sum_{j=1}^{n}{x_j}}{n},
\end{equation}

\begin{equation}
    S=\sqrt{\frac{\sum_{j=1}^{n}{x_j^2}}{n}-(\bar{X})^2}.
\end{equation}

Spatial clusters can be detected based on the $G_{i}^{*}$ value. The higher the value, the more significant the cluster is. However, the $G_{i}^{*}$ value just indicates whether there is a spatial cluster or not. To achieve our goal of pooling, we need to summarize the local region of the feature map by extracting a representative value. We use a threshold to do this. If the computed $G_{i}^{*}$ is greater than or equal to the threshold, a spatial cluster is detected and the value $x_i$ is used for pooling, otherwise the maximum in the window is used. 


\begin{equation}
   \mathcal{G}-pooling(\mathbf{x})=\begin{cases}
x_i& \text{if }G_{i}^{*} \geq \text{threshold}\\
max(\mathbf{x})& \text{if }G_i^* < \text{threshold}
\end{cases}
\end{equation}
It's noted that $G_{i}^{*}$ is in range [-2.8,2.8] where a negative value indicates a coldspot which means a spatial scatter and a positive value indicates a hotspot which means a spatial cluster. The absolute value $|G_{i}^{*}|$ indicates the significance. For example, a high positive $G_{i}^{*}$ value indicates the feature is more likely to be a spatial cluster.

The output feature map produced by $\mathcal{G}$-pooling is $\mathcal{G}$-pooling($\mathbf{X}$) which results after sliding the window over the entire input feature map. The threshold is set to 3 different values in this work, 1.0, 1.5, 2.0. A higher threshold means the current feature map has less chance to be reported as a spatial cluster and so max pooling will be applied instead. A lower threshold causes more spatial clusters to be detected and max pooling will be applied less often. As the threshold ranges from 1.0 to 1.5 to 2.0, fewer spatial clusters/hotspots will be detected. We find that a threshold of 2.0 results in few hostpots being detected and max pooling mostly to be used.


\begin{figure}
    \centering
      \includegraphics[width=0.95\linewidth]{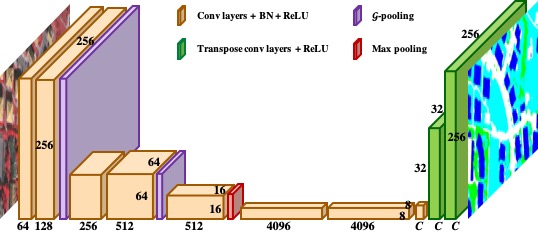}
      \caption{A FCN network architecture with $\mathcal{G}$-pooling.}
      \label{fig:framework}
\end{figure}

\subsection{Network Architecture}
\label{sec:architecture}

A pretrained VGG network \cite{simonyan2014vgg} is used in our experiments. VGG has been widely used as a backbone in various semantic segmentation networks such as FCN \cite{long2015fully}, U-net \cite{unet2015}, and SegNet\cite{segnet2017pami}. In VGG, the standard max pooling is a 2$\times$2 window size with a stride of 1. Our proposed $\mathcal{G}$-pooling uses a 4$\times$4 window size with a stride of 4. Therefore, after applying the standard pooling, the size of feature map drops to 1/2, while with our $\mathcal{G}$-pooling it drops to 1/4. A small window size is not used in our proposed $\mathcal{G}$-pooling since Getis-Ord $G_{i}^{*}$ analysis may not work well in such a small region. However, we tested the scenario where standard pooling is performed with a $4\times4$ sliding window and the performance is only slightly different from that using the standard $2\times2$ window. In general, segmentation networks using VGG16 as the backbone have 5 max pooling layers. So, when we replace max pooling with our proposed $\mathcal{G}$-pooling, there will be two $\mathcal{G}$-pooling and one max pooling layers.

\begin{table*}[htbp]
  \centering
  \caption{Experimental results of FCN using VGG-16 as backbone. Stride conv, $\mathcal{P}$-pooling and ours $\mathcal{G}$-pooling are used to replaced the standard max/average pooling.  }
    \begin{tabular}{l|ccccc|cc}
    \toprule
    \multicolumn{8}{l}{Potsdam} \\
    \midrule
        Methods & Roads & Buildings & Low Veg. & Trees & Cars  & mIoU  & Pixel Acc. \\
    \hline
    Max&70.62 & 74.28 & 65.94 & 61.36 & 61.40 & 66.72 & 79.55 \\
   Average& 69.34 & 74.49 & 63.94 & 60.06 & 60.28 & 65.62 & 78.08 \\
   Stride& 67.22 & 73.97 & 63.01 & 60.09 & 59.39 & 64.74 & 77.54 \\
    $\mathcal{P}$-pooling&\textbf{71.97} & 75.55 & 66.80 & 62.03 & 62.39 & 67.75 & 81.02 \\
    \hline
    $\mathcal{G}$-pooling-1.0 (ours) & 68.59 & \textbf{77.39} & 67.48 & 55.56 & 62.18 & 66.24 & 79.43 \\
    $\mathcal{G}$-pooling-1.5 (ours) &  70.06 & 76.12 & \textbf{67.67} & \textbf{62.12} & \textbf{63.91} & \textbf{67.98} & \textbf{81.63} \\
    $\mathcal{G}$-pooling-2.0 (ours) & 70.99 & 74.89 & 65.34 & 61.57 & 60.77 & 66.71 & 79.46 \\
    \midrule
    \multicolumn{8}{l}{Vaihingen} \\
    \midrule

      Max& 70.63 & 80.42 & 51.57 & 70.12 & 55.32 & 65.61 & 81.88 \\
    Average&70.54 & 79.86 & 50.49 & 69.18 & 54.83 & 64.98 & 79.98 \\
    Strde conv &68.36 & 77.65 & 49.21 & 67.34 & 53.29 & 63.17 & 79.44 \\
  $\mathcal{P}$-pooling &  71.06 & \textbf{80.52} & 51.70 & \textbf{70.93} & 53.65 & 65.57 & \textbf{82.44} \\
    \hline
    $\mathcal{G}$-pooling-1.0 (ours) & \textbf{72.15} & 79.69 & \textbf{53.28} & 70.89 & \textbf{53.72} & \textbf{65.95} & 81.78 \\
    $\mathcal{G}$-pooling-1.5 (ours) & 71.61 & 78.74 & 48.18 & 68.53 & 55.64 & 64.54 & 80.42 \\
    $\mathcal{G}$-pooling-2.0 (ours) & 71.09 & 78.88 & 50.62 & 68.32 & 54.01 & 64.58 & 80.75  \\
    \bottomrule
    \end{tabular}%
  \label{tab:results}%
\end{table*}%

\section{Experiments}
\subsection{Dataset}

\paragraph{ISPRS dataset}  We evaluate our method on two image datasets from the ISPRS 2D Semantic Labeling Challenge \cite{isprs_challenge}. These datasets are comprised of very high resolution aerial images over two cities in Germany: Vaihingen and Potsdam. While Vaihingen is a relatively small village with many detached buildings and small multi-story buildings, Potsdam is a typical historic city with large building blocks, narrow streets and dense settlement structure. The goal is to perform semantic labeling of the images using six common land cover classes: buildings, impervious surfaces (e.g. roads), low vegetation, trees, cars and clutter/background.  We report test metrics obtained on the held-out test images.

\noindent \textbf{Vaihingen}
The Vaihingen dataset has a resolution of 9 cm/pixel with tiles of approximately $2100\times2100$ pixels. There are 33 images, from which 16 have a public ground truth. Even though the tiles consist of Infrared-Red-Green (IRRG) images and DSM data extracted from the Lidar point clouds, we use only the IRRG images in our work. We select five images for validation (IDs: 11, 15, 28, 30 and 34) and the remaining 11 for training, following \cite{maggiori2017high,sherrah2016fully}.

\noindent \textbf{Potsdam}
The Potsdam dataset has a resolution of 5 cm/pixel with tiles of $6000\times6000$ pixels. There are 38 images, from which 24 have public ground truth. Similar to Vaihingen, we only use the IRRG images. We select seven images for validation (IDs: 2\_11, 2\_12, 4\_10, 5\_11, 6\_7, 7\_8 and 7\_10) and the remaining 17 for training, again following \cite{maggiori2017high,sherrah2016fully}.


\subsection{Experimental settings}
\paragraph{Baselines}Here, we compare our proposed $\mathcal{G}$-pooling with the standard max-pooling, average-pooling, stride convolution, and the recently proposed $\mathcal{P}$-pooling \cite{p-pooling2019cvpr}.  Max/average pooling is commonly for downsampling in the semantic segmentation networks that have VGG as a backbone. ResNet \cite{he2016resnet} is proposed without using any pooling but strided convolution. Such a network architecture has been adopted by recent studies for semantic segmentation, in particular the DeepLab series \cite{chen2018deeplabv3plus,chen2018deeplabv3,chen2017deeplabv2} and PSPNet\cite{zhao2017pspnet}. Max pooling is removed and instead strided convolution is used to downsample the feature maps while dilated convolution is used to enlarge the receptive fields. There is also work on detail preserving pooling, for example DDP \cite{saeedan2018ddp} and $\mathcal{P}$-pooling \cite{p-pooling2019cvpr}. We select the most recent one, $\mathcal{P}$-pooling, which outperforms the other detail preserving methods for comparison.

\subsection{Evaluation Metrics}
We have two goals in this work, the model's segmentation accuracy and its generalization performance. Model accuracy is used to report the performance on the test/validation set using the model trained with training set within one dataset. Model generalizability is used to report the performance of the test/validation set with another dataset. In general, the domain gap between train and test/validation set from one dataset is relatively small. However, cross-dataset testing exists large domain shift problem. 
\vspace{-10pt}
\paragraph{Model accuracy} The commonly used per class intersection over union (IoU) and mean IoU (mIoU) as well as the pixel accuracy are adopted for evaluating segmentation accuracy.
\vspace{-10pt}
\paragraph{Model generalizability} Specifically, we will perform evaluation on the ISPRS Potsdam set with a model trained on the ISPRS Vahingen set (Potsdam$\rightarrow$Vaihingen) and reverse the order (Vaihingen$\rightarrow$Potsdam). Pixel accuracy and mIoU are used to report the performance of the generalizability.

\begin{table*}[t]
  \centering
  \caption{Cross-location evaluation. We compare the generalization capability of using $\mathcal{G}$-pooling with domain adaptation method AdaptSegNet which utilize the unlabeled data.}
    \begin{tabular}{l|ccccc|cc}
    \toprule
       
    \multicolumn{8}{c}{Potsdam $\rightarrow$ Vaihingen} \\
    \midrule
          & Roads & Buildings & Low Veg. & Trees & Cars  & mIoU  & Pixel Acc. \\
    \hline
    Max-pooling & 28.75 & 51.10 & 13.48 & \textbf{56.00} & 25.99 & 35.06 & 47.48 \\
    stride conv &  28.66 & 50.98 & 12.76 & 55.02 & 24.81 & 34.45 & 46.51 \\
$\mathcal{P}$-pooling & 32.87 & 50.43 & 13.04 & 55.41 & 25.60 & 35.47 & 48.94 \\
    Ours ($\mathcal{G}$-pooling) & \textbf{37.27} & \textbf{54.53} & \textbf{14.85} & 54.24 & \textbf{27.35} & \textbf{37.65} & \textbf{55.20} \\
    \hline
    AdaptSegNet & 41.54 & 40.74 & 21.68 & 50.45 & 36.87 & 38.26 &  57.73\\
     \hline
     \hline
    \multicolumn{8}{c}{Vaihingen $\rightarrow$ Potsdam} \\
    \hline
    
    Max-pooling & 20.36 & 24.51 & 19.19 & 9.71  & 3.65  & 15.48 & 45.32 \\
    stride conv &  20.65 & 23.22 & 16.57 & 8.73  & 8.32  & 15.50 & 42.28 \\
     $\mathcal{P}$-pooling & 23.97 & 27.66 & 14.03 & \textbf{10.30} & 12.07 & 19.61 & 44.98 \\
    Ours ($\mathcal{G}$-pooling) & \textbf{27.05} & \textbf{29.34} & \textbf{33.57} & 9.12  & \textbf{16.01} & \textbf{23.02} & \textbf{45.54} \\
    \hline
    AdaptSegNet & 40.28 & 37.97 & 46.11 & 15.87 & 20.16 & 32.08 & 50.28 \\

    \bottomrule
    \end{tabular}%
  \label{tab:cross-test}%
\end{table*}%
\subsection{Implementation Details}
\paragraph{Implementation of $\mathcal{G}$-pooling} Models are implemented using the PyTorch framework. Max-pooling, average-pooling, stride conv are provided as built-in function and $\mathcal{P}$-pooling has open-source code. We implement our $\mathcal{G}$-pooling in C and use the interface to connect to PyTorch for network training. We adopt the network architecture of FCN \cite{long2015fully} with a backbone of a pretrained VGG-16 \cite{simonyan2014vgg}. The details of the FCN using our $\mathcal{G}$-pooling can be found in Section \ref{sec:architecture}. The results in Table \ref{tab:results} are reported using FCN with a VGG-16 backbone. 
\vspace{-10pt}
\paragraph{Training settings} Since the image tiles are too large to be fed through a deep CNN due to limited GPU memory, we randomly extract image patches of size of 256$\times$256 pixels as the training set. Following standard practice, we only use horizontal and vertical flipping as data augmentation during training. For testing, the whole image is split into $256\times256$ patches with a stride of 256. Then, the predictions of all patches are concatenated for evaluation.

 We train all our models using Stochastic Gradient Descent (SGD) with an initial learning rate of 0.1, a momentum of 0.9, a weight decay of 0.0005 and a batch size of 5. If the validation loss plateaus for 3 consecutive epochs, we divide the learning rate by 10. If the validation loss plateaus for 6 consecutive epochs or the learning rate is less than 1e-8, we stop the model training. We use a single TITAN V GPU for training and testing.

\begin{table}[htbp]
  \centering
  \caption{The average percentage of detected spatial clusters per feature map with different threshold. }
    \begin{tabular}{lccc}
    \toprule
    Threshold & 1.0 & 1.5 & 2.0 \\
    \midrule
    Potsdam & 15.87 & 9.85  & 7.65 \\
    Vaihingen & 14.99 & 10.44 & 7.91 \\
    \bottomrule
    \end{tabular}%
  \label{tab:threshold}%
\end{table}%

\section{Effectiveness of $\mathcal{G}$-pooling}
\label{sec:pooling_results}

In this section, we first show that incorporating geospatial knowledge into a pooling function of the standard CNN learning can improve segmentation accuracy. Then we demonstrate the promising generalization capability of our proposed $\mathcal{G}$-pooling. 

\begin{figure*}
    \centering
    \includegraphics[width=\linewidth]{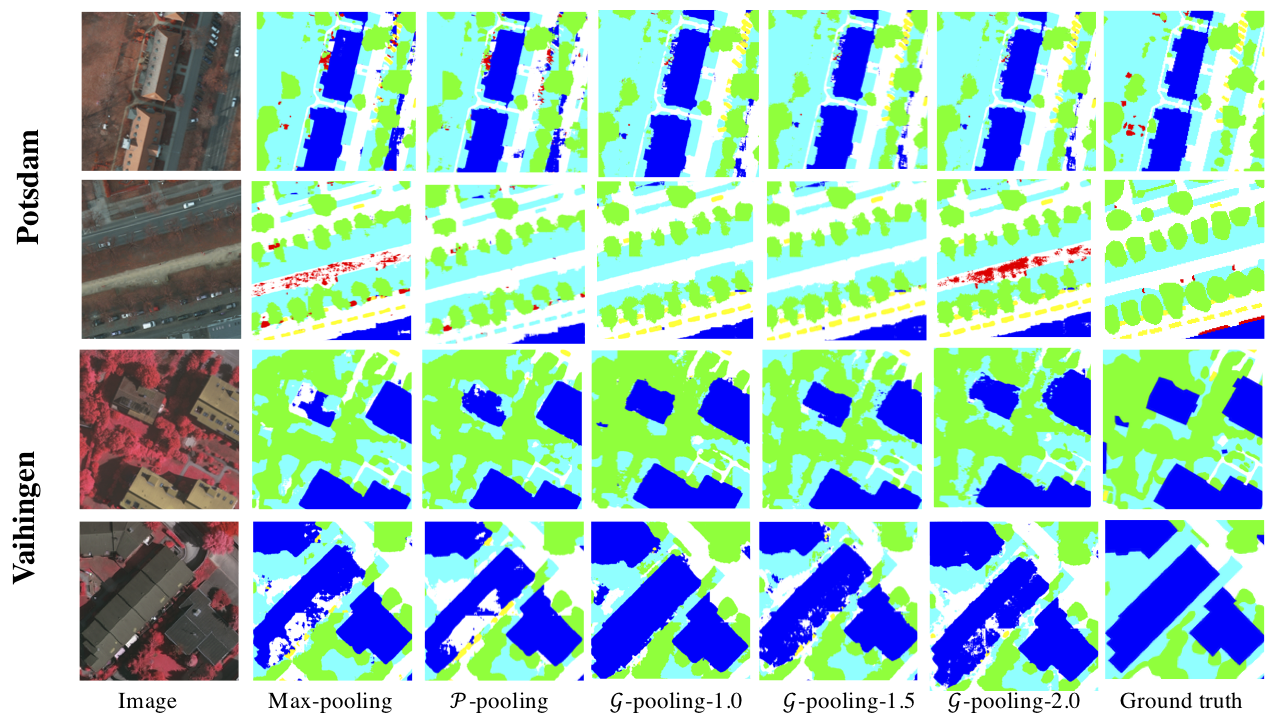}
    \caption{Qualitative results of ISPRS Potsdam. White: road, blue: building, cyan: low vegetation, green: trees, yellow: cars, red: clutter.}
    \label{fig:results}
\end{figure*}

The segmentation accuracy on FCN using various pooling functions reported on the test set is shown in Table \ref{tab:results}. For $\mathcal{G}$-pooling, we experiment on 3 different thresholds, which is 1.0, 1.5 and 2.0. The range of $G_{i}^{*}$ value is [-2.8, 2.8]. As explained in Section \ref{sec:defination}, higher $G_{i}^{*}$ value can cause more uses of max pooling. If we set the $G_{i}^{*}$ value as 2.8, then the case will be all max pooling.  Qualitative results are shown in Figure \ref{fig:results}. And the quantitative results for evaluating model accuracy and cross-location generalization is shown in Table \ref{tab:results} and \ref{tab:cross-test} respectively.


 \vspace{-10pt}

\paragraph{Non-spatial vs geospatial statistics} The baselines of pooling functions are usually non-spatial statistics, for example, finding the max/average value. Our approach provides a geospatial process to simulate how things are related based on spatial location. Here, we pose the question, ``\textit{is the knowledge useful to train a deep CNN?}''.  As we mentioned in Section \ref{sec:methods}, such a knowledge incorporated method can bring the benefit of improved generalizability. As shown in Table \ref{tab:results}, for Potsdam, using geospatial knowledge to design the pooling function can bring 1.23\% improvement compared to $\mathcal{P}$-pooling. Our $\mathcal{G}$-pooling-1.0 and 2.0 is not able to outperform some baselines in the model accuracy testing, which indicates the threshold selection is important. Some classes of the baselines have higher performance compared to ours. This is expected since the dataset is relatively small and may be overfitting. The qualitative results in Figure \ref{fig:results} show our proposed $\mathcal{G}$-pooling has less pepper-and-salt effect. In particular, there is less noise inside the objects compared to the other methods. This demonstrates our proposed $\mathcal{G}$-pooling simulates the geospatial distributions and makes the prediction within the objects more compact. The effects of threshold is shown in Table \ref{tab:threshold}, as described in Section \ref{sec:methods}, the higher the threshold the less spatial cluster detected.

\vspace{-10pt}
\paragraph{Domain adaptation vs knowledge incorporation}  Table \ref{tab:cross-test}
compares using pooling functions with using unsupervised domain adaptation (UDA). We note that the UDA method AdaptSegNet \cite{tsai2018learning} uses a large amount of unlabeled data from the target dataset to adapt the model which has been demonstrated to help generalization. The other methods don't benefit from the unlabeled data. As shown in Table \ref{tab:cross-test}, our proposed $\mathcal{G}$-pooling is able to achieve the best generalization performance. For Potsdam$\rightarrow$Vaihingen, $\mathcal{G}$-pooling outperforms $\mathcal{P}$-pooling by more than 2\%. For Vaihingen$\rightarrow$Potsdam, the improvement is even more significant, at least 3.41\%. When we compare the knowledge incorporation method $\mathcal{G}$-pooling with the domain adaptation method AdaptSegNet, the performance difference is just 0.61\% for Potsdam. The results verify our assumption that incorporating knowledge helps generalizations as well. And the performance is close to that of domain adaptation which utilizes a great amount of unlabeled data to learn the data distribution. Even though knowledge incorporation doesn't outperform data-based domain adaptation, these two methods can be combined to provide even better generalization.

\section{$\mathcal{G}$-pooling and state-of-the-art methods}
\label{sec:segmentation_results}
In order to verify that our proposed $\mathcal{G}$-pooling is able to improve state-of-the-art segmentation approaches, we select DeepLab \cite{chen2017deeplabv2} and SegNet \cite{segnet2017pami} as additional network architectures to test $\mathcal{G}$-pooling. As mentioned above, the models in Section \ref{sec:pooling_results} use FCN as the network architecture and VGG-16 as the backbone. For fair comparison with FCN, VGG-16 is also used as the backbone in DeepLab and SegNet.

DeepLab \cite{chen2017deeplabv2} uses a large receptive fields through dilated convolution. For the baseline DeepLab itself, \textit{pool4} and \textit{pool5} from the backbone VGG-16 are removed and followed by \cite{tsai2018learning} and the dilated conv layers with a dilation rate of 2 are replaced with $conv5$ layers. For the $\mathcal{G}$-pooling version, \textit{pool1,pool2} are replaced with $\mathcal{G}$-pooling and we keep \textit{pool3}. Thus there are three max pooling layers in the baseline and one $\mathcal{G}$-pooling layer and one max pooling layer in our proposed version.  SegNet uses an encoder-decoder architecture and preserves the max pooling index for unpooling in the decoder. Similar to Deeplab, there are 5 max pooling layers in total in the encoder of SegNet so \textit{pool1,pool2} are replaced with the proposed  \textit{$\mathcal{G}$\_pool1} and \textit{pool3,pool4} are replaced with \textit{$\mathcal{G}$\_pool2}, and \textit{pool5} is kept. This leads us to use a $4\times4$ unpooling window to recover the spatial resolution where the original ones are just $2\times2$. Thus there are two $\mathcal{G}$-pooling and one max pooling layers in our SegNet version. 

As can be seen in Table \ref{tab:model}, $\mathcal{G}$-pooling is able to improve the model accuracy for Potsdam, 67.97\% $\rightarrow$ 68.33\%. And the improvement on the generalization test Potsdam$\rightarrow$Vaihingen is even more obvious, $\mathcal{G}$-pooling improves mIoU from 38.57 to 40.04. Similar observations can be made for SegNet and FCN. For Vaihingen, even though the model accuracy is not as high as the baseline, the difference is small. The mIoU of our versions of DeepLab, SegNet and FCN is less than 1\% lower. We note that Vaihingen is an easier dataset than Potsdam, since it only includes urban scenes while Potsdam includes both urban and nonurban. However, the generalizability of our model using $\mathcal{G}$-pooling is much better. As shown, when testing Potsdam using a model trained on Vaihingen, FCN with $\mathcal{G}$-pooling is able to achieve 23.02\% mIoU which is an improvement of 7.54\% IoU. The same observations can be made for DeepLab and SegNet.

\begin{table}[htbp]
  \centering
  \caption{Experimental results on comparing w/o and w/ proposed $\mathcal{G}$-pooling for the state-of-the-art segmentation networks. P$\rightarrow$V indicates the model trained on Potsdam and test on Vaihingen, and versa the verses.}
    \begin{tabular}{l|c|cc|cc}
    \toprule
          & \multicolumn{3}{c|}{Potsdam (P)} & \multicolumn{2}{c}{P$\rightarrow$V} \\
    \hline
    Network & $\mathcal{G}$-Pooling & mIoU  & PA & mIoU  & PA \\
    \hline
    \multirow{2}[2]{*}{DeepLab} & $\times$      & 67.97            &81.25   &38.57           & 58.47 \\
          & \checkmark                          & 68.33   &80.67            &\textbf{40.04}  & \textbf{63.21} \\
    \hline
    \multirow{2}[2]{*}{SegNet} &  $\times$      &69.47            &82.53            &35.98           &53.69\\
          & \checkmark                          &70.17   &83.27   &\textbf{39.04}  & \textbf{56.42} \\
    \hline
    \multirow{2}[2]{*}{FCN} &  $\times$         &66.72             & 79.55           & 35.06          & 47.48 \\
          & \checkmark                          &67.98    & 81.63  & \textbf{37.65} & \textbf{55.20}  \\
    \hline
          & \multicolumn{3}{c|}{Vaihingen (V)} & \multicolumn{2}{c}{V$\rightarrow$P} \\
    \hline
    \multirow{2}[2]{*}{DeepLab} &  $\times$      & 70.80   & 83.74    & 18.44           &33.96 \\
          & \checkmark                           & 70.11            & 83.09             & \textbf{19.26}  &\textbf{36.17} \\ 
    \hline
    \multirow{2}[2]{*}{SegNet} &  $\times$      & 66.04             & 81.79              &16.77           &45.90 \\
          & \checkmark                          & 66.71    & 82.66     &\textbf{25.64}  &\textbf{48.08}\\
    \hline
    \multirow{2}[2]{*}{FCN} &  $\times$         & 65.61              &81.88    &15.48           & 45.32 \\
          & \checkmark                          & 65.95     & 81.87             &\textbf{23.02}  & \textbf{45.54}\\
    \bottomrule
    \end{tabular}%
  \label{tab:model}%
\end{table}%

\section{Discussion}


Incorporating knowledge is not a novel approach for neural networks. Before deep learning, there was work on rule-based neural networks which required expert knowledge to design the network for specific applications. Due to the large capacity of deep models, deep learning has become the primary approach to address vision problems. However, deep learning is a data-driven approach which relies significantly on the amount of training data. If the model is trained with a large amount of data then it will have good generalization. But the case is often, particularly in overhead image segmentation, that the dataset is not large enough like it is in ImageNet/Cityscapes. This causes overfitting. Early stopping, cross-validation, etc. can help to avoid overfitting. Still, if domain shift exists between the training and test sets, the deep models do not perform well. In this work, we propose a knowledge-incorporated approach to reduce overfitting. We address the question of how to incorporate the knowledge directly into the deep models by proposing a novel pooling method for overhead image segmentation. But some issues still need discussing as follows.

\paragraph{Scenarios using $\mathcal{G}$-pooling} As mentioned in section \ref{sec:methods}, $\mathcal{G}$-pooling is developed using Getis-Ord $G_{i}^{*}$ analysis which quantifies how the spatial convergence occurs. This is a simulated process design for geospatial data downsampling. Thus it's not necessarily appropriate for other image datasets. This is more general restriction of incorporating of knowledge. The Getis-Ord $G_{i}^{*}$ provides a method to identify spatial clusters while training. The effect is similar to conditional random fields/Markov random fields in standard computer vision post-processing methods. However, it is different from them since the spatial clustering is dynamically changing based on the feature maps and the geospatial location while post-processing methods rely on the prediction of the models.

\paragraph{Local geospatial pattern}
We now explain how $\mathcal{G}$-pooling works in deep neural networks. Getis-Ord $G_{i}^{*}$ analysis is usually used to analyze a global region hotspot detection which describes the geospatial convergence. As shown in Figure \ref{fig:framework}, $\mathcal{G}$-pooling will be applied twice to downsample the feature map. The spatial size of the $\mathcal{G}$-pooling will be $64\times64$ and $16\times16$ respectively. And the max-pooling will lead to the size of feature map being reduced by $1/2$ while ours it will be by $1/4$. This is because we want to compute $G_{i}^{*}$ over a larger region.

Even though $G_{i}^{*}$ is usually computed over a larger region than in our framework, it still provides captures spatial convergence within a small region. Also, two $\mathcal{G}$-pooling operations are applied at different scales of feature map and so a larger region in the input image is really considered. Specifically, the first $4\times4$ pooling window is slid over the $256\times256$ feature map and the output feature map has size $64\times64$. This is fed through the next conv layers and a second $\mathcal{G}$-pooling is applied. At this stage, the input feature map is $64\times64$ and so when a $4\times4$ sliding window is now used, a region of $16\times16$ is really considered, which is 1/16 of the whole image.

\paragraph{Limitations} There are some limitations of our work. For example, we didn't investigate the optimal window size for performing Getis-Ord $G_{i}^{*}$ analysis. We also only consider one kind of spatial pattern, clusters. And, there might be better places than pooling to incorporate knowledge in CNN architectures. 

\label{sec:results}

\section{Conclusion}
In this paper, we investigate how geospatial knowledge can be incorporated into deep learning for geospatial image analysis. We demonstrate that incorporating geospatial rules improves performance. We realize, though, that ours is just preliminary work into geospatial guided deep learning. We note the limitations of our approach, for example, that the prior distribution does not provide benefits for classes in which this prior knowledge is not relevant. Our proposed approach does not show much improvement on the single dataset case especially a small dataset. ISPRS Vaihingen is a very small dataset which contains around only 500 images of size of $256\times 256$. In the future, we will explore other ways to encode geographic rules so they can be incorporated into deep learning models.



{\small
\bibliographystyle{ieee_fullname}
\bibliography{egbib}
}

\end{document}